\documentclass[runningheads]{llncs}
\usepackage{graphicx}
\usepackage{algorithm}
\usepackage{amsmath,amssymb} 

\usepackage{color}
\usepackage[width=122mm,left=12mm,paperwidth=146mm,height=193mm,top=12mm,paperheight=217mm]{geometry}
\usepackage{wrapfig,blindtext}
\usepackage{capt-of}

\usepackage{array}
\newcolumntype{P}[1]{>{\centering\arraybackslash}p{#1}}

\usepackage{wrapfig}
\usepackage{subfig}

\usepackage{breqn}
\usepackage[font=footnotesize, labelfont={sf}, margin=1cm]{caption}

\begin{document}
\pagestyle{headings}
\mainmatter

\title{Cross-modal Supervision for Learning Active Speaker Detection in Video}

\author{Punarjay Chakravarty and Tinne Tuytelaars}
\institute{KU Leuven, ESAT-PSI-iMinds}

\maketitle

\begin{abstract}
In this paper, we show how to use audio to supervise the learning of active speaker detection in video. Voice Activity Detection (VAD) guides the learning of the vision-based classifier in a weakly supervised manner. The classifier uses spatio-temporal features to encode upper body motion - facial expressions and gesticulations associated with speaking. We further improve a generic model for active speaker detection by learning person specific models. Finally, we demonstrate the online adaptation of generic models learnt on one dataset, to previously unseen people in a new dataset, again using audio (VAD) for weak supervision. The use of temporal continuity overcomes the lack of clean training data. We are the first to present an 
active speaker detection system that learns on one audio-visual dataset and automatically adapts to speakers in a new dataset. This work can be seen as an example of how the availability of multi-modal data allows us to learn a model without the need for supervision, by transferring knowledge from one modality to another.
\end{abstract}

\section{Introduction}

The problem of detecting active speakers in video is a central one to several applications. In video conferencing, knowing the active speaker allows the application to focus on and transmit the video of one amongst several people at a table. In a Human-Computer-Interaction (HCI) setting, a robot/computer can use active speaker information to address the correct interlocuter. Active speaker detection is also a part of the pipeline in video diarization, the automatic annotation of speakers, their speech and actions in video. Video diarization is useful for movie sub-titling, multimedia retrieval and for video understanding in general.

Traditionally, visual active speaker detection has been done using lip motion detection ~\cite{elKhoury14,Everingham06,Everingham09,Haider12}. However, facial expressions and gestures from the upper body, movement of the hands, etc., are all cues that can be utilized to assist with this task,
as shown in ~\cite{Chakravarty15}, where better detection results are achieved using spatio-temporal features extracted from the entire upper body, compared with just lip motion detection.

Another powerful idea we borrow from~\cite{Chakravarty15}, is to use audio to supervise the training of a video based active speaker detection system.

In that work, a microphone array is used to get directional sound information (assumed to be speech sounds), and based on this input, upper body tracks are associated with speak/non-speak labels. These labels are then used to train an active speaker classifier using video only.

\begin{wrapfigure}[21]{r}{0.5\textwidth}
\centering
\vspace*{-1.1cm} 
    \includegraphics[width=0.48\textwidth,height=0.35\textwidth]{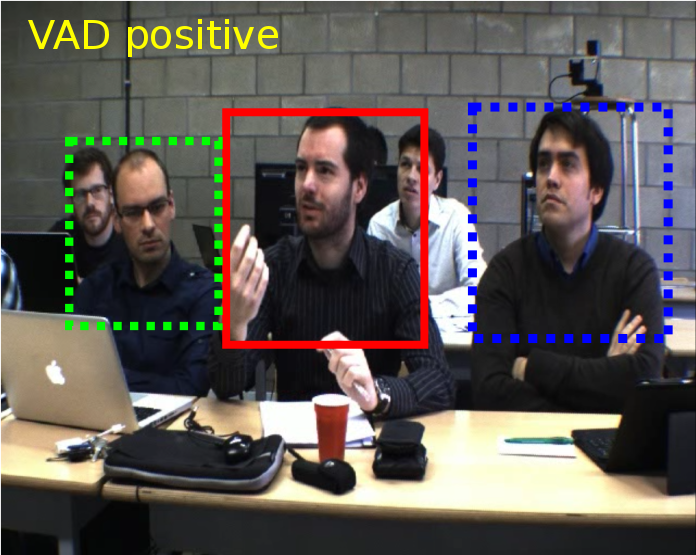} 
   \caption{Audio-based Voice Activity Detection (VAD) is used to weakly supervise the training of a video-based active speaker classifier. VAD tells us that someone in the frame is speaking, but not who. The problem is one of associating the voice activity with one of the people (solid red upper body bounding box) in the frame, and training the classifier at the same time. We use structured output learning to train a latent SVM classifier in the presence of partially observed (latent) inputs.}
\label{fig:problemStatement}
\end{wrapfigure}

However, the presence of reverberation and background noise prevents perfect active speaker identification using directional audio alone, which subsequently affects the training of the video-based classifier.
Additionally, in the vast majority of videos, such as the millions of Youtube videos available online, in videos from films and TV series, only a single channel of sound is available, with no directional information. Even in those cases where 2 channels of audio are available,  the relative position of the camera and the microphones varies, and no calibration information is available, making it impossible to apply the method of~\cite{Chakravarty15}.

In the absence of directional information, we propose to use Voice Activity Detection (VAD)~\cite{Germain13} to tell us when there is someone speaking in a frame. If there is only one person in the frame, then this can be used to train the video-based classifier directly. However, when this is not the case, the problem becomes one of simultaneously associating the voice activity with one of the people in the frame, and learning the classifier (Figure \ref{fig:problemStatement}). That's the challenge we address in this work.

Moreover, there's an additional challenge. Investigating our trained classifier, we find that it has some bias: it works better for some speakers, compared to others. We identify two reasons for this. First, the way people gesticulate while speaking varies a lot from person to person. Indeed, a person-specific model typically
outperforms the generic model. 
Second, there is the domain shift problem: the change of data distribution between training and test data. We address both by extending our previous scheme to an online learning setting that, starting from a generic classifier, gradually adapts to a specific person. To this end, we retrain the model with an incrementally increasing number of training samples coming from a new video of a previously unseen person at each iteration. The online training is also weakly supervised by VAD from audio. The generic classifier is used to label and pick the training samples for each speaker and temporal continuity constraints allow the classification performance to improve in spite of imperfectly labelled training data from the generic classifier.



Our method is completely unsupervised, in the sense that there is no \emph{human} supervision/labelling. We use audio to supervise the learning. This supervision comes "for free" with the video, but is only partial - VAD tells us that one of the persons in the frame is speaking, but not who. As opposed to~\cite{Chakravarty15}, who use full supervision from directional audio, we use weak supervision from VAD.
This work can be seen as an example of how the availability of multi-modal data allows us to learn a model without the need for supervision, by transferring knowledge from one modality to another.

The remainder of the paper is organized as follows.
We discuss prior work in this area in Section \ref{relWork}. We discuss the use of audio for active speaker detection in Section \ref{audioTraining}, with subsections \ref{latentSVM}, \ref{speakerSpecificModels} and \ref{onlineLearning} discussing the weakly supervised learning with Latent SVMs, speaker specific classification and online learning, respectively. Experimental results are discussed in  Section \ref{expts} and concluding remarks and potential for future work in Section \ref{concl}.

\section{Related Work}
\label{relWork}

\paragraph{Weakly supervised and multimodal learning} 
The learning of a classifier in the presence of weak supervision, or partially labelled data, has been studied mostly in the context of object recognition, where labels are available for images, but localization information - bounding boxes around the objects to be classified, are missing ~\cite{Bilen14a,Bilen14,Bilen15,Deselaers12,Song14}. 
Best results in this context are obtained with Structured Output Learning ~\cite{Nguyen09}, i.e. by learning a classifier that outputs not only the class labels, but also the bounding box coordinates or index.
We use the same approach for training a classifier for active speaker detection with only VAD-based supervision, which gives us labels for the images, but not for individual bounding boxes. Audio weakly supervises the training of video.
The work of Bojanowski et al. ~\cite{Bojanowski13} is another example of one mode of information weakly supervising another. They use scripts to weakly supervise the learning of actors and actions in movies. 
However, scripts are not always available for video data, while audio is.

\paragraph{Dealing with domain shift}
In our work, we find that an active speaker classifier trained on a first set of speakers performs less well on previously unseen speakers, while best results are obtained with person-specific classifiers. This is because of the mismatch between the distributions of different speakers.
On the one hand, training a generic classifier means that it has seen a larger number of training samples, is less prone to overfitting compared to a person specific classifier, and should generalize well for unseen speakers. However, the generic classifier still suffers from person-specific biases, and gives better classification results for some people over others. 
The same problem exists for object recognition - a classifier trained on one dataset typically has lower performance when applied to images from another dataset.
This is known as the dataset bias problem, and there have been some efforts at reducing this for object recognition ~\cite{Khosla12,Tommasi13}.
One way to deal with person or dataset specific biases is to adapt the source classifier to the target classifier, and this is called Domain Adaptation ~\cite{Aljundi15,Fernando13}.
Transfer Learning~\cite{Aytar11,Tommasi09,Tommasi10}, a related problem, is about using the information available from the source data to aid the learning of the target classifier utilizing only a small number of target training samples.
For instance, Aytar et al.~\cite{Aytar11} use an Adaptive SVM (ASVM) that incrementally adapts an SVM learnt on source data (e.g. motorbike class) to target data (e.g. bicycle class) in the context of object recognition. The source classifier acts as a regularizer for the target classifier in the adaptive SVM framework, and they demonstrate successful adaptation based on a relatively small number of training samples of the target class.
This work lies at the basis of our online adaptation to previously unseen persons.

\paragraph{Person-specific models} There has been some work on person specific facial expression recognition and transferring generic to specific models for improving classification performance~\cite{Chen12,Chu13,Zen14}.
Chen et al.~\cite{Chen12} show that facial expression recognition results improve when using person specific classifiers. They use an Inductive Transfer Learning (ITL) approach, where they learn a source classifier, which is a collection of weak learners in a boosting framework. Subsequently a subset of these are used for training the target classifier with a small number of labeled target samples. 

Chu et al.~\cite{Chu13} propose a Selective Transfer Machine (STM) approach to re-weight the source samples so that they are closer to the target samples. The algorithm simultaneously learns the parameters of the classifier and the source sample weights that minimize the error between the source and target distributions. They thus personalize a generic classifier to individual, with the resulting personalized classifier improving on the generic classifier on facial action unit detection tasks.
However, STM requires the storage of all source samples, with a higher memory requirement than storing just the source classifier, which could be the weights of an SVM.

Zen et al.~\cite{Zen14} demonstrate unsupervised adaptation of a generic classifier to a target classifier on single frame expression datasets. They learn a regression function between the ``shape'' or sample distribution of each user in the labelled source dataset and his/her classifier (source weight vector $w_i$ in the SVM). 
Applying this function on the unlabelled sample distribution of the target user then gives them the target classifier (target weight vector $w_t$). They do not require to keep in memory all the samples from the source dataset and outperform the STM method of ~\cite{Chu13}. However, their approach requires that the relative distribution of positive and negative samples in every user's data is relatively constant 
and can be learnt using the source users. However, this is not the case in our data. Additionally, we learn the generic source classifier using unlabelled data as well - so our process requires no human supervision from beginning to end.

\paragraph{Online learning} 
is the incremental learning of a classifier with an increasing number of training samples as and when they become available. In our context, we adapt the generic source classifier to the person-specific target classifier with an increasing number of samples from the speaker. This is somewhat similar to the problem of Active Learning, where a new classifier is to be learnt with the minimum budget in terms of time spent in labelling training samples, and the task is one of selecting the most relevant samples to be used for training. Gavves et al.~\cite{Gavves15} demonstrate Active Transfer Learning, in that the selection of relevant training samples is done with the help of previously learnt classifiers on other datasets. Both~\cite{Gavves15} and~\cite{Zen14} use the source classifiers as zero-shot priors, giving a baseline performance using only the target classifier, with classification performance gradually increasing with an increasing number of samples from the target dataset. We use this as our inspiration for our online learning problem, except again, our learning is without any manual supervision.

\section{Audio Supervised Training}
\label{audioTraining}

In the original experiment of~\cite{Chakravarty15}, a 2-mic array was used to associate upper bodies detected in the video, with sound directions.
They used a technique proposed by ~\cite{Sayeh14} for estimating the number and direction of sound sources. A non-linear function of the Generalized Cross Correlation Phase Transform (GCC-PHAT) between the audio signals is calculated over all the angles of arrival with respect to the microphone array baseline. This is done over short time intervals corresponding to the Time Frequency cells of a Short Term Fourier Transform. This gives an angle of arrival spectrum at each point in time that can be associated with the people detected in the image.
In each frame, the sound direction is associated with a speaker's upper body bounding box, and features within that bounding box are used to train the classifier.

We use the same data as ~\cite{Chakravarty15}, available on request from the authors, and consider the case when directional information is absent. We simulate the output of VAD by removing the speak/non-speak bounding box labels. We assign a label of speak to the frame if any of the bounding boxes in it are tagged as speaking and non-speak if none of the bounding boxes is speaking.

Our problem is one of associating one of the bounding boxes in the image with the sound and training a classifier at the same time.
We treat this  as a structured output prediction problem ~\cite{Pletscher10}.

\subsection{Classifier Training Under Weak Audio Supervision using Structured Output Learning}
\label{latentSVM}

In the absence of information about which upper body bounding box is associated
with the active speaker in each frame, the problem can be posed as a structured
training problem~\cite{Bilen14a,Bilen14,Bilen15,Nguyen09}, in the presence of partially observed training data. 

In the context of object recognition, there are databases with images labelled with the presence of one or more objects in the scene, but no localization (bounding box) information for the object in the image. ~\cite{Bilen14a,Bilen14,Bilen15} deal with this by using a Latent SVM formulation, which alternates between the guessing of object bounding boxes, and training a classifier for the object inside the bounding box. They use object proposals ~\cite{Uijlings13} to narrow down the search for objects in the image. 

Here, we adapt~\cite{Bilen14a,Bilen14,Bilen15} to our setting. Our object proposals are the upper body bounding boxes. We know that one of the bounding boxes is an active speaker, but not which one - the speak/non-speak label for the individual bounding boxes are our latent variables. 
Using structured output prediction, we jointly learn which of the bounding boxes in the image is associated with the active speaker, together with learning the active speaker classifier.

Given an image $x$ and upper body bounding box $h$, let $\phi(x,h)$ denote an image description computed over bounding box $h$. 
Given all upper body bounding boxes $h_1,...h_n$, the algorithm then needs to select the bounding box that contains the active speaker.
The labels of the images, speaking/non-speaking, $y= \pm 1$, are obtained from the sound using VAD or, in our experiment, by removing the directional information from the training data.
Once the classifier is trained, the best bounding box $h$ is found by 
\begin{equation}
h^* = \underset{h}{\operatorname{argmax}}  \langle w,\phi(x,h) \rangle
\label{eq:bestBB}
\end{equation}
where $w$ is the weights vector of the SVM.
We define $\Phi(x,y,h) = \phi(x,h)$ if $y = 1$, and $0$ otherwise.
The learning task is to optimize the following:
\begin{equation}
\hat{w} = \underset{w}{\operatorname{argmin}} \sum_{i=1}^{N} l(w,x^i,y^i) + \frac{C}{2} \|w\| ^2
\end{equation}
where $l(w,x^i,y^i)$ is the per example loss, $\frac{C}{2} \|w\| ^2$ is the regularizer and $N$ is the total number of training data.
The max-margin loss function is defined as 
\begin{dmath}
l_{mm}(w,x^i,y^i) = \underset{y,h}{\operatorname{max}} ( \langle w,\Phi(x^i,y,h) \rangle +
\Delta(y^i,y)) - \underset{h}{\operatorname{max}} ( \langle w,\Phi(x^i,y^i,h) \rangle )
\end{dmath}
where $\Delta(y^i,y)$ is the zero-one error, which is $0$ if $y^i=y$ and 1 otherwise.

This loss function tries to maximize the margin between the score of the selected active speaker's bounding box and the non-speaking bounding boxes.
Following the work of~\cite{Bilen14} and~\cite{Bilen15}, we replace the max-margin loss with a soft-max loss function:
\begin{dmath}
l_{sm}(w,x^i,y^i) = \frac{1}{\beta} log \sum_{y,h} exp(\beta \langle w, \Phi(x^i,y,h) \rangle + \beta \Delta(y^i,y)) - \frac{1}{\beta} log \sum_{h} exp(\beta \langle w, \Phi(x^i,y^i,h) \rangle)
\end{dmath}

where $\beta$ controls the sharpness of the distribution. It can be shown that as $\beta \rightarrow \infty$, the loss function limits to the standard structured SVM formulation.
The softmax loss function allows for multiple active speakers in the same frame. It also
makes the optimization function smoother and less prone to local minima. We use the LBFGS solver from minFunc \footnote{\url{http://people.cs.ubc.ca/~schmidtm/Software/minFunc.html}
} to optimize our cost function and train our classifier.


\subsection{Speaker Specific Models}
\label{speakerSpecificModels}

Using the motion of the face and upper body over time assists with active speaker detection. At the same time, it maybe has the disadvantage of making the detector more speaker specific, as different people are likely to have different mannerisms while speaking. We explore this hypothesis by training several person specific Active Speaker classifiers.
We do this in two settings: one using the directional audio (i.e., supervised), as a baseline, and subsequently, in the VAD setting, where the learning is weakly supervised by audio, as detailed in the previous section.

In the first case, the learning is straightforward: we have a separate track for each person in the video, and knowledge of the frames in which that track is speaking (from the directional audio). 

In the second case, the audio does not tell us which track/person is speaking at any given time, just that one among the multiple tracks in the frame is speaking. 
For this, we do the training in two steps. We first learn a generic classifier in the weakly supervised case, as detailed previously. Subsequently, we use the generic (source) classifier to guide the selection of the positive samples for the person specific (target) classifier. We run the generic classifier on each ``speaking'' frame's bounding boxes to get an idea of which track/bounding box is speaking.
However, the generic classifier does not always give the highest score to the active speaker in the frame. This is because of the dataset bias and domain shift problem discussed earlier - the generic classifier performs better for some speakers compared to others. So we bring in another cue: temporal continuity.

So far, we have discussed active speaker detection on each frame in isolation. However, people's speech tends to be for periods longer than a single frame. If a person is speaking in one frame, it is more likely than not, that they will be speaking in the next frame as well. 

We use temporal continuity to reduce the effect of mis-classifications of the generic classifier and guide the sample selection for the speaker specific classifier. 
The highest scoring sample at each VAD-positive frame is taken to be the positive sample for the associated speaker, and all other samples are selected as negative samples for the other speakers. Both positive and negative samples are weighted according to temporal continuity, measured as the number of contiguous neighbouring frames with consistent labels.
We use a weighted logistic loss function $l_{wll}$

\begin{dmath}
l_{wll}(w,\Phi(x,y,h^*),\alpha) = \alpha \cdot log\lbrace1+exp(-\langle w,\Phi(x,y,h^*) \rangle )\rbrace
\label{eq:weightedLogLoss}
\end{dmath}

where $\Phi(x,y,h^*)$ is the feature vector from the best scoring bounding box, $w$ is the weights vector of the speaker-specific SVM and $\alpha$ is the temporal continuity weight of the sample.

Note how this integration of temporal continuity directly in the weakly supervised learning framework (as opposed to keeping it as a postprocessing step, as is usually done) reflects again one of the core ideas behind our work, that combining multiple, independent sources of information - be it multiple modalities, or temporal vs. spatial information - allows learning models with less supervision. 


\subsection{Online Learning}
\label{onlineLearning}
In this section, we deal with the problem of learning the specific model in an online fashion for a speaker who has not been seen earlier during training. This can be the case during a live setting, where we don't have the entire data available to us at any given time, just what we have seen so far. To this end, we use a model inspired by the Tabula Rasa Transfer Learning model of Aytar et. al.~\cite{Aytar11}.

The idea is that the generic model is used as a zero-shot prior, and already gives a baseline performance, that can be improved as a new speaker specific model is trained incrementally with every additional batch of samples that trickle in from the new speaker. This allows us to have a model that performs better than the prior, generic model in an iterative fashion, without needing to see all the target samples. 
The process of online learning of speaker-specific classifiers is again weakly supervised by audio: it assumes that VAD is available for the target speaker data as well.

As in the offline case for training speaker specific models (subsection \ref{speakerSpecificModels}), we use VAD to detect the frames in which human speech is present. Subsequently, the generic (source) classifier is used to guide the selection of the positive samples for each new speaker (target). We select the highest scoring bounding box in each VAD-positive frame as the positive sample for the speaker associated with it, and the remaining bounding boxes are selected as negative samples for the other speakers. Temporal continuity is used to weigh both the positive and negative samples (Equation ~\ref{eq:weightedLogLoss}).

Motivated by~\cite{Gavves15}, we use the prior (source) model, not just for the selection of the target speaker's positive training samples, but also for target prediction.
During prediction, the generic model scores are added to the target model scores 
so that the prediction score from online learning, at each iteration is given as:

\begin{dmath}
f^{t}(\phi(x,h))=\langle w^{gen}, \phi(x,h) \rangle + \langle w^{t}, \phi(x,h) \rangle
\label{eq:predictionScore}
\end{dmath}

Each time step $t$ has an increasing number of training samples to train the classifier $w^{t}$ at that iteration. $ w^{gen}$ remains constant during online learning.

This results in the person-specific target classifier being at least as good as the generic source classifier, and getting progressively better with an increasing number of training samples.



\section{Experiments}
\label{expts}

We use the audio-visual dataset made available by the authors of ~\cite{Chakravarty15}. It consists of 7 recordings of masters student thesis presentations to a jury of examiners. Each student presents for 25 minutes, followed by 5 minutes of questioning by the jury. The microphone array, with its directional sound information in a cone of 180 degrees in front of it is associated with upper body tracks of the jury. We will call this the Masters student dataset in the rest of the paper. An example frame of this data is shown in Figure ~\ref{fig:problemStatement}.

\cite{Chakravarty15} used the directional sound information from the microphone array, associated with the bounding boxes of persons in the frame to train their video-based active speaker classifier. We simulate VAD by removing this directional information from the data, leaving only a label of speak/non-speak per frame.

\begin{figure}[t]
\begin{center}
    \includegraphics[width=0.80\linewidth]{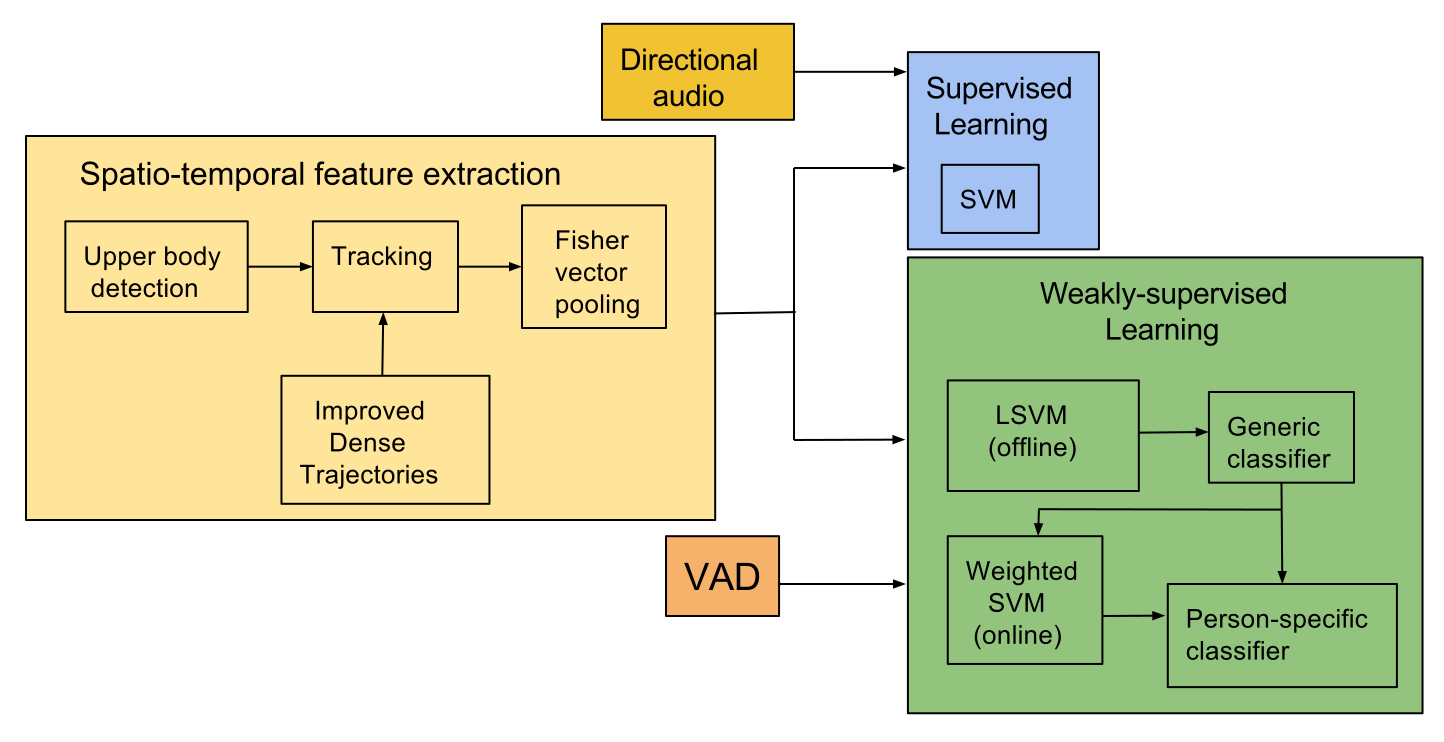} 
\end{center}
 \vspace*{-0.6cm} 
   \caption{Experimental setup}
\label{fig:systemDiagram}
 \vspace*{-0.6cm} 
\end{figure}

Like \cite{Chakravarty15}, we only use the 3 people from the jury in the front row of the audience, as others behind them are obscured. The people in all the experiments are the same, and do not change positions. 

We train the active speaker detection classifier in a Leave-One-Out-Cross-Validation (LOOCV) fashion, where the data from 6 presentations are used for training, and tested on the 7th presentation. This is repeated 7 times. 

Finally, we test the model learnt on the Masters dataset on an entirely new dataset that we present - the Columbia dataset. It is an 87-minute-long video of a panel discussion at Columbia university, available from YouTube \footnote{\url{https://youtu.be/6GzxbrO0DHM}}. There are 7 speakers on the panel, and the camera focusses on smaller groups of speakers at a time. We only focus on the parts of the video where there is more than one person in the frame, and ignore people on the margins of the video who are not detected by the upper body detector. This gives us sections of video for 5 speakers, with 2-3 speakers visible at any one time. We have annotated the upper body bounding boxes of each speaker with speak/non-speak labels, about 35 minutes of video in all, which will be made available. 

We update the generic classifier learnt on the Masters dataset online, in a completely unsupervised fashion, with the generic classifier adapting to each new speaker in the Columbia dataset, with subsequent improvement in performance.

\subsection{Implementation Details}

We use the same improved trajectory features (ITF)~\cite{Wang13} recommended by~\cite{Chakravarty15}, for training our active speaker detection classifier. ITF are spatio-temporal features used for state of the art action recognition, and comprise of a concatenation of Histogram of Oriented Gradients (HOG), Histogram of Flow (HoF) and Motion Boundary Histogram (MBH) features. HoG, HoF and MBH features are calculated in the immediate neighbourhood of each point on the grid. We use 15 consecutive frames for calculating the ITF - this corresponds to about 7 seconds of video in the Masters dataset. The HoG, HoF and MBH features are independently reduced to half their original dimensions using PCA, and feature vectors from within an upper body track are pooled using  Fisher vectors (FV)~\cite{Perronnin10}.

We apply intra-class L2 normalization, power and a final L2 normalization of the whole FV before  classification using a linear SVM. We use a codebook size of 256 for the FV encoding. The FV encoding is done independently for HOG, HoF and MBH, before they are concatenated to a single, 101,376 dimensional vector. Intra-class L2 normalization - normalization within each block of the FV related to a single codeword, is used to balance weights of the different codewords in the FV, and  reduces the "burstiness" in the FVs (often resulting from features belonging to the background). Training a linear SVM with a non-linear feature map (obtained using the power normalization) has the advantage of approximating a non-linear SVM at lower computational complexity ~\cite{Vedaldi12}.
These techniques, recommended as best practice in ~\cite{Peng14}, have been to shown to considerably boost performance of FVs. 

For upper body detection, we use a detector trained using the Deformable Parts Model from \cite{voc-release5}. The tracking is relatively straight-forward, because people don't change positions and there are no crossing tracks.

ITF are grouped by their start frame (calculated from the following 15 frames), and a FV is calculated for all the improved trajectories within a bounding box (person) track starting from that frame.

The active speaker classifier is sensitive to the frame-rate of the dataset on which it is trained. To have the classifier transfer between datasets, we subsample the Columbia dataset so that its frame-rate matches the frame-rate of the Masters dataset.

A pipeline of the system is shown in Figure
\ref{fig:systemDiagram}.

\begin{wraptable}[9]{r}{0.5\textwidth}
\centering
 \vspace*{-0.75cm} 
\begin{tabular}
{|P{1.7cm}|P{2.0cm}|P{2.0cm}|}
\hline
 & Directional Audio & VAD \\
      \hline
Avg. AUC & 0.69 \tiny$\pm 0.07$  & 0.71 \tiny$\pm0.05$ \\
\hline
\end{tabular}
\caption{Average AUC (with standard deviations) for active speaker detection fully supervised by directional audio \cite{Chakravarty15}, and weakly supervised by VAD, over all experimental folds (Masters dataset).}
\label{tab:directionalvsVAD}
 \vspace*{-0.7cm} 
\end{wraptable}

\subsection{Weak Supervision Using Audio}

VAD results in frames with speak/non-speak labels. There are no speak/non-speak labels for individual bounding boxes and the FVs extracted from them. 
Section \ref{latentSVM} details the Structured Output SVM classifier that is used for training the active speaker detection classifier in the absence of training labels for individual bounding boxes.

Table \ref{tab:directionalvsVAD} displays the results of our 
experiments with the active speaker detection classifier trained using VAD. The results with weak supervision (structured output learning) are comparable with the results from fully supervised learning from directional sound. This shows that the structured output formulation and the soft-max loss function for optimization transfers well from the object localization application of~\cite{Bilen14,Bilen15}, to our task of active speaker localization in the absence of bounding box labels for training.


\subsection{Speaker Specific Models}
\label{speakerSpecificModelsExpts}
Section \ref{speakerSpecificModels} makes the hypothesis that training person specific active speaker detection models will give better results than training a generic model for all speakers. To validate this hypothesis, we perform three experiments: 

\begin{enumerate}
\item Full directional audio (giving speak/non-speak labels for all bounding boxes in the frame) for training the person specific classifier.

\item VAD audio (speak/non-speak label for the frame, but without information about individual bounding boxes) for training the person-specific classifier. This highest scoring sample using the generic classifier is used to get positive training samples for each person in a VAD-positive frame.

\item Experiment 2, with samples weighted according to temporal continuity (see Eq.~\ref{eq:weightedLogLoss}).

\end{enumerate}

\begin{wraptable}[10]{r}{0.60\textwidth}
\vspace*{-1.1cm} 
\centering
\small
\begin{tabular}
{|P{0.8cm}|P{1.4cm}|P{1.4cm}|P{1.4cm}|P{1.8cm}|}
\hline
 Expt.  &  Speaker 1 &  Speaker 2 &  Speaker 3 &  Mean Avg. AUC\\ 
      \hline
1 & 0.79 \tiny$\pm0.08$ & 0.76 \tiny$\pm0.03$ & 0.88 \tiny$\pm0.05$ & 0.81 \tiny$\pm0.07$ \\ 
\hline
2 & 0.60 \tiny$\pm0.10$ & 0.59 \tiny$\pm0.07$ & 0.75 \tiny$\pm0.03$ & 0.65 \tiny$\pm0.10$ \\ 
\hline
3 & 0.79 \tiny$\pm0.10$ & 0.80 \tiny$\pm0.03$ & 0.88 \tiny$\pm0.04$ & 0.82 \tiny$\pm0.07$ \\ 
\hline
\end{tabular}
\caption{Mean Avg AUC (with standard deviations) for person-specific active speaker detection using (1) directional audio, (2) VAD - no temporal weighting \& (3) VAD with temporally weighted samples (All expts on Masters dataset).}
\label{tab:speak_specific}
\vspace*{-1.6cm} 
\end{wraptable}

When full directional audio supervision is available (expt. 1), the speaker specific models show better results, a 10\% improvement over the generic classifier of Table~\ref{tab:directionalvsVAD}. 


When using VAD for weak supervision with a hard-max posterior (expt 2), the person-specific classifier performs worse (16\% worse mean average AUC) than the person-specific classifier with full audio supervision (expt 1), and worse even than the generic classifier. 

This confirms the dataset bias problem we discussed in Section ~\ref{relWork}. The generic classifier might be more biased towards one speaker compared to the others and occasionally score the true positive speaker lower than another non-speaker in the same VAD-positive frame. This leads to the use of mis-classified samples for the training of the person-specific classifiers in the weakly supervised case, and their subsequent poor performance.

In experiment 3, a temporal weight is added to each sample - the number of contiguous neighbouring frames in which it has been consistently labelled  (see Equation \ref{eq:weightedLogLoss}). We use a temporal window of 3 seconds. This results in a mean average AUC of 0.82, comparable to the fully supervised case (expt 1). This shows that the temporal weighting of samples correctly guides the sample selection. Thus, it acts as another weak supervisor (apart from the VAD) for the training of the speaker specific classifer. Table \ref{tab:speak_specific} presents results for all 3 speaker-specific  experiments in the Masters dataset.
It should be noted here that for all experiments in this sub-section, the evaluations are performed on individual frames and temporal continuity is exploited as an extra cue during training, not as a postprocessing step to correct results afterwards.

\subsection{Online Learning}
Here, we report results of experiments that demonstrate how a generic classifier trained on speakers in the Masters dataset, can be modified online, to specific speakers in the Columbia dataset.

We only select sections of video in which there are 2 or more people in the frame at the same time. This is to demonstrate the unsupervised selection of training samples from one among many speakers. The selection of training samples when only 1 speaker is present in the frame is trivial (VAD can be used to detect positive and negative samples for the speaker), and is not considered in this experiment.

The prior classifier is run on each VAD-positive frame in the new dataset and the highest scoring bounding box is taken to be the positive sample for that speaker in the frame, and the remaining bounding boxes are taken to be the negative samples for the other speakers. This assumes that there is only one person speaking at a time in the video, which is actually the case in most target applications. The samples are weighted according to their temporal continuity - a positive sample with a higher number of contiguous positive samples around it gets a higher weight, as was done in number 3 of the speaker specific experiments (subsection ~\ref{speakerSpecificModelsExpts}). 

The experiment begins by using the prior classifier to detect active speakers in the new data. Then, with each iteration of online learning, a balanced selection of positive and negative samples are selected from each speaker, and used for training the person-specific classifier. The number of training samples increases with each iteration.

\begin{figure}[t]
\vspace*{-0.4cm} 
  \centering
  \subfloat[]{\includegraphics[width=6.1cm,height=5cm]{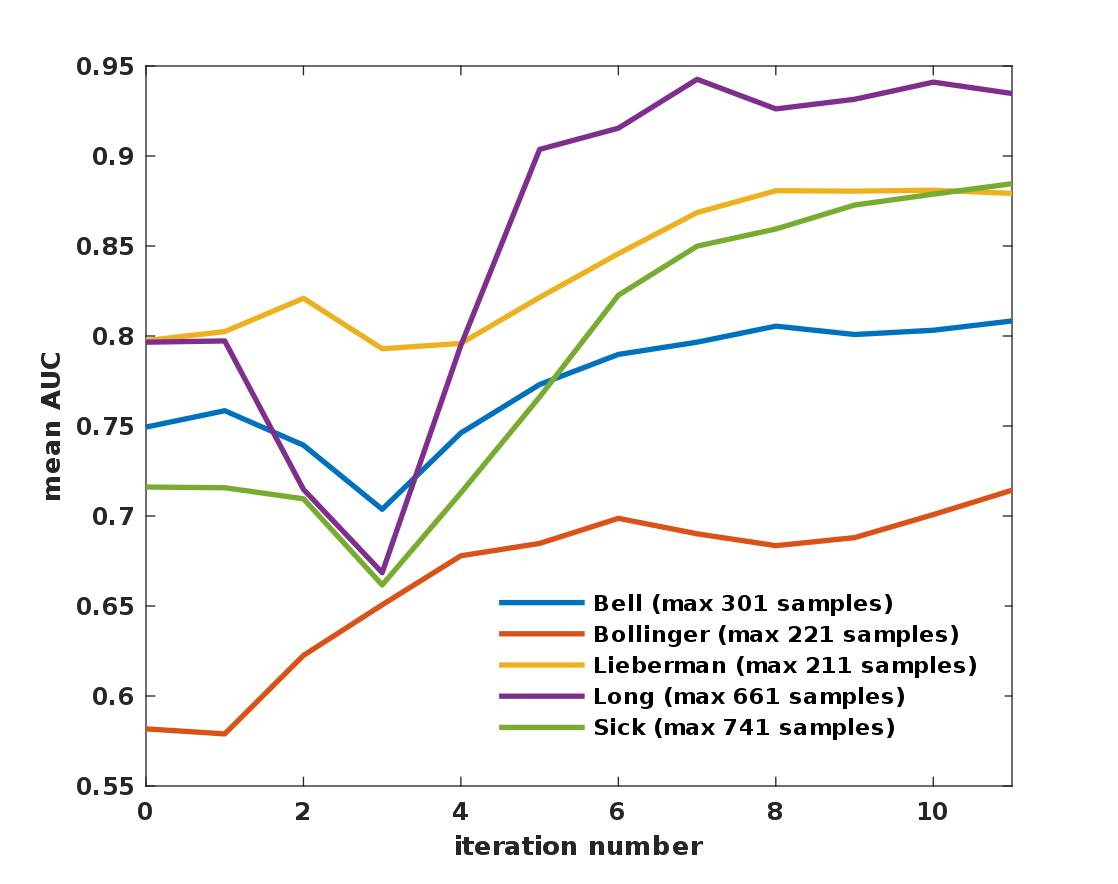}\label{fig:online_learning}}
  \hfill
  \subfloat[]{\includegraphics[width=6.1cm,height=5cm]{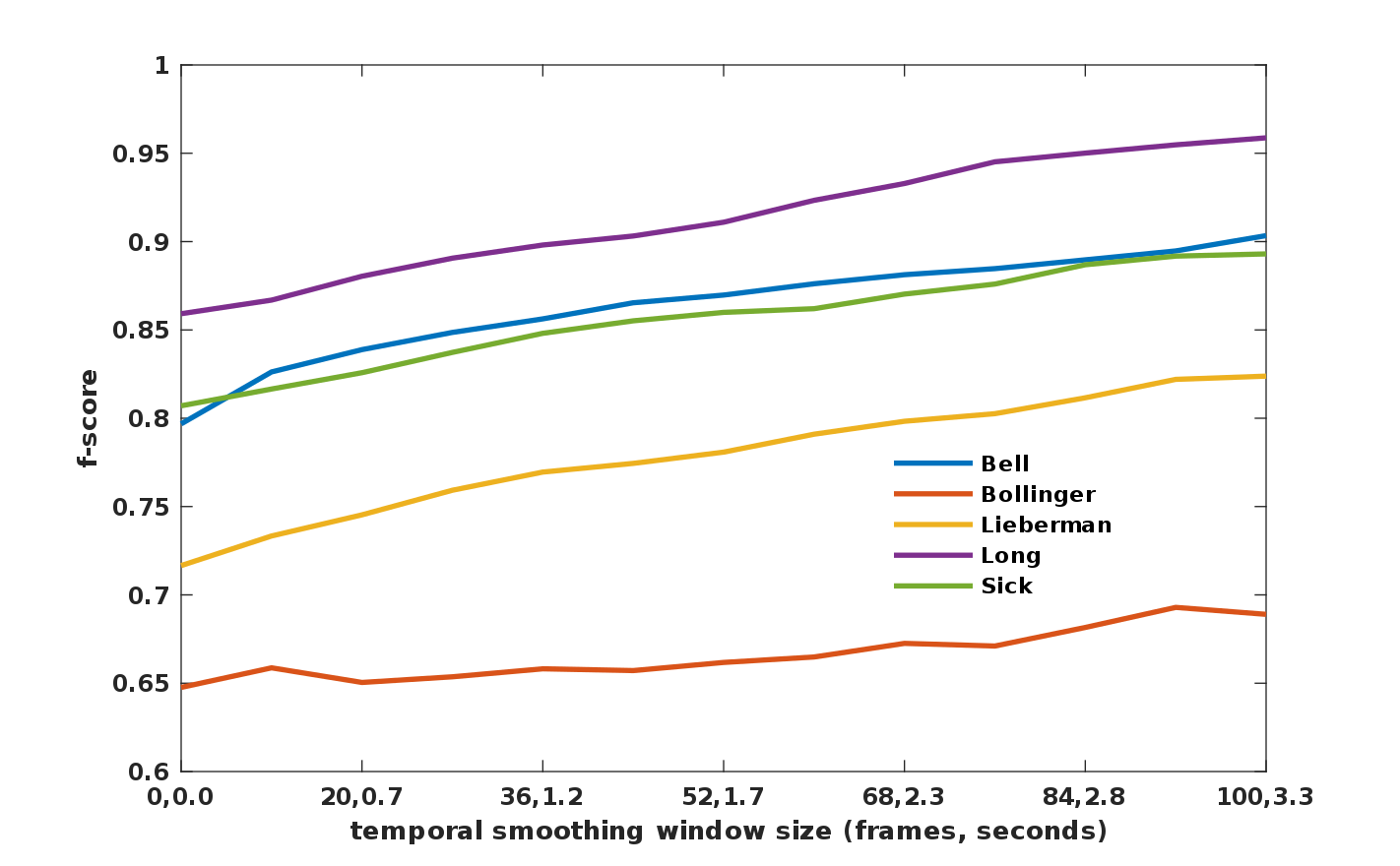}\label{fig:temporal_smoothing}}
     \caption{(a) Online Learning: Mean AUC over all speakers in the new Columbia dataset with an increasing number of training samples in each iteration. (b) Temporal smoothing: F-scores for all speakers at the end of online learning, after thresholding and temporal smoothing, with increasing size of temporal window in the Columbia datset.}
\vspace*{-0.8cm} 
\end{figure}

Figure \ref{fig:online_learning} displays the mean average AUC results for experiments conducted per speaker over the training iterations. We see that the performance of the iteratively trained person specific target classifier starts out at the performance of the generic source classifier, and gradually improves with increasing number of target training samples.

There is an initial dip in the performance of the classifier learnt online for 3 of the 5 people, when there is a small number of training samples. If some of these samples are wrongly selected by the prior classifier, 
then the classifier's performance will decrease to a level below the generic classifier performance. But, as the speaker speaks for longer, and more correct samples weighted by their temporal continuity are picked, the online learning adapts to the target distribution. 

We use a maximum of 10 seconds of video per person for the online learning in the Columbia dataset in our experiments, and see an improvement of about 5-15\% over the performance of the prior classifier. Thus, our method of selecting samples weighted by their temporal continuity is resilient against the selection of some wrong samples, and very quickly - within a few seconds - adapts to each new speaker.

We use temporal continuity to further improve the performance of the online-learnt classifier during inference as well. The scores from the classifier learnt during the last iteration of online learning are thresholded (at the intersection of the ROC curve with the diagonal) and smoothed over increasing lengths of time (from 0 to about 3 seconds). Figure \ref{fig:temporal_smoothing} shows that the f-scores for all the speakers improve with increasing amounts of temporal smoothing, with plateauing of results at around 3 seconds. A potential downside of too much smoothing is that if a person speaks for short durations (single, yes/no utterances for example), then these are not going to be registered. The amount of temporal smoothing applied would depend on the application. For video conferencing, it might not be appropriate to switch focus between speakers for such short utterances, and a smoothing of 3 seconds (the maximum smoothing applied in our experiments during inference), would probably be adequate.

\begin{figure}[t]
\begin{center}
    \includegraphics[width=0.90\linewidth]{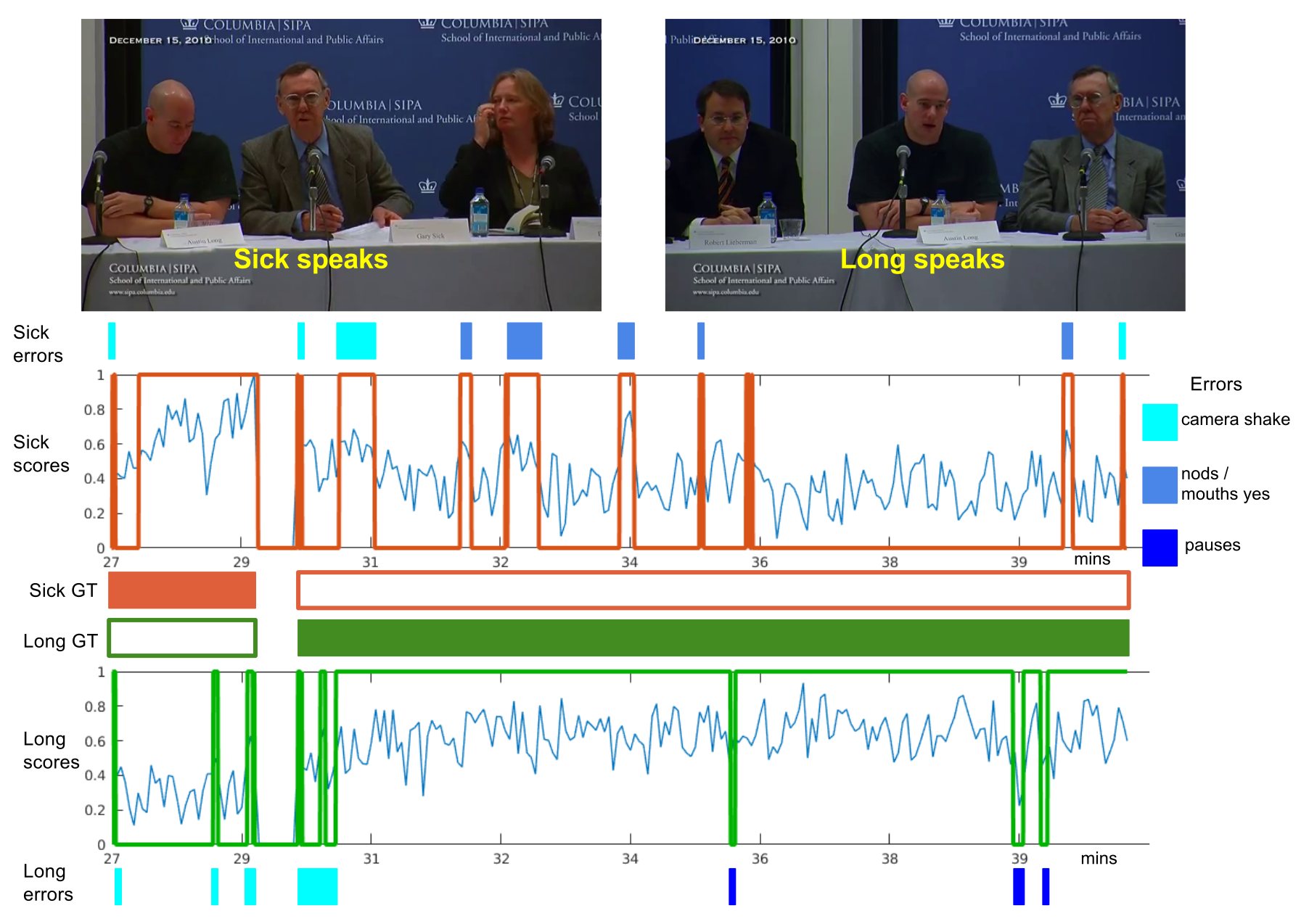} 
\end{center}
\vspace*{-0.6cm} 
   \caption{Normalized raw scores (blue) with the online-learnt classifier and thresholded and temporally smoothed speak/non-speak values for speakers Sick (red) and Long (green), along with Ground Truth (GT, solid colour = speak), in minutes 27:00 to 40:00 in the Columbia dataset. Temporal smoothing is done over 96 frames (about 3 seconds of video).}
\label{fig:activeSpeakerDetectionErrors}
 \vspace*{-0.6cm} 
\end{figure}

Figure ~\ref{fig:activeSpeakerDetectionErrors} shows a timeline for Active Speaker Detection in the Columbia dataset, for speakers Sick and Long, during minutes 27:00 to 40:00 in the video. The classifiers for these speakers are learnt online earlier in the video, and the raw scores for these speakers over time are shown in blue. The scores are thresholded and temporally smoothed to obtain speak/non-speak values, shown in red and green for Sick and Long respectively. Ground truth speak/non-speak values for these speakers are also given. It can be seen that the parts of the video where the algorithm apparently makes a mistake can be explained by camera shake, or where a non-active speaker actually nods and mouths yes in response to another active speaker (ground truth does not mark this as speech), or when an active speaker pauses mid-sentence.

\section{Conclusions}
\label{concl}
This paper demonstrates the use of audio for cross-modal supervision of the training of a video-based active speaker detector. The problem is posed in terms of a structured output prediction problem - given information about the presence of an active speaker in a frame from audio-based Voice Activity Detection, find out which particular person is speaking, among the people in the frame, and at the same time, learn the video-based classifier for active speaker detection. Person-specific classifiers are shown to perform better than generic classifiers, and the learning of the specific classifiers is again weakly supervised by audio. 
The prior classifier adapts to the specific speaker using samples from just a few seconds of video, with additional improvement in results using temporal smoothing. This shows that the system has the potential to be used in a video conferencing application, and quickly learn the characteristics of new speakers.

In future work, we will close the loop between audio and video. In current work, audio supervises the learning of a video-based person-specific active speaker detector. The learnt video classifier will in turn supervise the learning of person-specific voice models and those voice models will be fed back into the video to further improve active speaker detection. This is expected to be particularly useful in the more challenging data encountered in video diarization: movies and TV series with non-frontal views of people, where the video-only classifier is expected to perform worse than in frontal-view video.

\clearpage

\bibliographystyle{splncs}
\bibliography{egbib}{}

\end{document}